\documentclass[english]{article}
\usepackage[margin=1.5in]{geometry}
\usepackage[T1]{fontenc}
\usepackage[latin9]{inputenc}
\usepackage{array}
\usepackage{verbatim}
\usepackage{rotfloat}
\usepackage{booktabs}
\usepackage{units}
\usepackage{url}
\usepackage{multirow}
\usepackage{amsmath}
\usepackage{amssymb}
\usepackage{graphicx}
\usepackage{footnote}
\makesavenoteenv{tabular}
\makesavenoteenv{table}

\usepackage[authoryear]{natbib}

\makeatletter

\title{Low Latency Privacy Preserving Inference}

\providecommand{\tabularnewline}{\\}

\usepackage{times}
\usepackage{titlesec}

\titleformat{\subsubsection}[runin]
  {\normalfont\normalsize\bfseries}{\thesubsubsection}{1em}{}

\makeatother

\usepackage{babel}
\begin{document}
\global\long\def\Z{\mathbb{Z}}

\global\long\def\xx{\mathbf{x}}

\global\long\def\ww{\mathbf{w}}

\global\long\def\vv{\mathbf{v}}

\global\long\def\mm{\mathbf{m}}

\global\long\def\cc{\mathbf{c}}

\global\long\def\rr{\mathbf{r}}

\global\long\def\ring{\mathcal{R}}

\global\long\def\EE{\mathbb{E}}

\global\long\def\DD{\mathbb{D}}

\author{Alon Brutzkus \\
	Microsoft Research and Tel Aviv University \\
	\texttt{alonbrutzkus@mail.tau.ac.il} \\
	\and
	Oren Elisha \\
	Microsoft \\
	\texttt{orelisha@microsoft.com} \\
	\and
	Ran Gilad-Bachrach \\
	Microsoft Research \\
	\texttt{rang@microsoft.com}
}

\maketitle
\begin{abstract}
When applying machine learning to sensitive data, one has to find
a balance between accuracy, information security, and computational-complexity.
Recent studies combined Homomorphic Encryption with neural networks
to make inferences while protecting against information leakage. However,
these methods are limited by the width and depth of neural networks
that can be used (and hence the accuracy) and exhibit high latency
even for relatively simple networks. In this study we provide two
solutions that address these limitations. In the first solution, we
present more than $10\times$ improvement in latency and enable inference
on wider networks compared to prior attempts with the same level of
security. The improved performance is achieved by novel methods to
represent the data during the computation. In the second solution,
we apply the method of transfer learning to provide private inference
services using deep networks with latency of $\sim0.16$ seconds.
We demonstrate the efficacy of our methods on several computer vision
tasks.
\end{abstract}

\section{Introduction}

Machine learning is used in several domains such as education, health,
and finance in which data may be private or confidential. Therefore,
machine learning algorithms should preserve privacy while making accurate
predictions. The privacy requirement pertains to all sub-tasks of
the learning process, such as training and inference. For the purpose
of this study, we focus on private neural-networks inference. In this
problem, popularized by the work on CryptoNets \citep{gilad2016cryptonets},
the goal is to build an inference service that can make predictions
on private data. To achieve this goal, the data is encrypted before
it is sent to the prediction service, which should be capable of operating
on the encrypted data without having access to the raw data. Several
cryptology technologies have been proposed for this task, including
Secure Multi-Party Computation (MPC) \citep{yao1982protocols,goldreich1987play},
hardware enclaves, such as Intel\textquoteright s Software Guard Extensions
(SGX) \citep{mckeen2013innovative}, Homomorphic Encryption (HE) \citep{Gen09},
and combinations of these techniques.

The different approaches present different trade-offs in terms of
computation, accuracy, and security. HE presents the most stringent
security model. The security assumption relies on the hardness of
solving a mathematical problem for which there are no known efficient
algorithms, even by quantum computers \citep{Gen09,albrecht2018homomorphic}.
Other techniques, such as MPC and SGX make additional assumptions
and therefore provide weaker protection to the data \citep{yao1982protocols,mckeen2013innovative,chen2018sgxpectre,koruyeh2018spectre}.

While HE provides the highest level of security it is also limited
in the kind of operations it allows and the complexity of these operations
(see Section~\ref{subsec:Homomrphic-Encryption}). CryptoNets \citep{gilad2016cryptonets}
was the first demonstration of a privacy preserving Encrypted Prediction
as a Service (EPaaS) solution \citep{sanyal2018tapas} based on HE.
CryptoNets are capable of making predictions with accuracy of $99\%$
on the MNIST task \citep{lecun2010mnist} with a throughput of $\sim59000$
predictions per hour. However, CryptoNets have several limitations.
The first is latency - it takes CryptoNets $205$ seconds to process
a single prediction request. The second is the width of the network
that can be used for inference. The encoding scheme used by CryptoNets,
which encodes each node in the network as a separate message, can
create a memory bottleneck when applied to networks with many nodes.
For example, CryptoNets' approach requires $100$'s of Gigabytes of
RAM to evaluate a standard convolutional network on CIFAR-10. The
third limitation is the depth of the network that can be evaluated.
Each layer in the network adds more sequential multiplication operations
which result in increased noise levels and message size growth (see
Section~\ref{sec:Applying-Deep-Nets}). This makes private inference
on deep networks, such as AlexNet \citep{krizhevsky2012imagenet},
infeasible with CryptoNets' approach.

In this study, we present two solutions that address these limitations.
The first solution is Low-Latency CryptoNets (LoLa), which can evaluate
the same neural network used by CryptoNets in as little as $2.2$
seconds. Most of the speedup ($11.2\times$ ) comes from novel ways
to represent the data when applying neural-networks using HE.\footnote{The rest of the speedup ($8.2\times$) is due to the use of a faster
	implementation of HE.} In a nut-shell, CryptoNets represent each node in the neural network
as a separate message for encryption, while LoLa encrypts entire layers
and changes representations of the data throughout the computation.
This speedup is achieved while maintaining the same accuracy in predictions
and 128 bits of security \citep{albrecht2018homomorphic}.\footnote{The HE scheme used by CryptoNets was found to have some weaknesses
	that are not present in the HE scheme used here.}

LoLa provides another significant benefit over CryptoNets. It substantially
reduces the amount of memory required to process the encrypted messages
and allows for inferences on wider networks. We demonstrate that in
an experiment conducted on the CIFAR-10 dataset, for which CryptoNets'
approach fails to execute since it requires $100$'s of Gigabytes
of RAM. In contrast, LoLa, can make predictions in $12$ minutes using
only a few Gigabytes of RAM.

The experiment on CIFAR demonstrates that LoLa can handle larger networks
than CryptoNets. However, the performance still degrades considerably
when applied to deeper networks. To handle such networks we propose
another solution which is based on transfer learning. In this approach,
data is first processed to form ``deep features'' that have higher
semantic meaning. These deep features are encrypted and sent to the
service provider for private evaluation. We discuss the pros and cons
of this approach in Section~\ref{sec:Applying-Deep-Nets} and show
that it can make private predictions on the CalTech-101 dataset in
$0.16$ seconds using a model that has class balanced accuracy of
$81.6\%$. To the best of our knowledge, we are the first to propose
using transfer learning in private neural network inference with HE.
Our code is freely available at https://github.com/microsoft/CryptoNets.

\section{Related Work}

The task of private predictions has gained increasing attention in
recent years. \citet{gilad2016cryptonets} presented CryptoNets which
demonstrated the feasibility of private neural networks predictions
using HE. CryptoNets are capable of making predictions with high throughput
but are limited in both the size of the network they can support and
the latency per prediction. \citet{bourse2017fast} used a different
HE scheme that allows fast bootstrapping which results in only linear
penalty for additional layers in the network. However, it is slower
per operation and therefore, the results they presented are significantly
less accurate (see Table~\ref{tab:MNIST-performance-comparison}).
\citet{boemer2018ngraph} proposed an HE based extension to the Intel
nGraph compiler. They use similar techniques to CryptoNets \citep{gilad2016cryptonets}
with a different underlying encryption scheme, HEAAN \citep{cheon2017homomorphic}.
Their solution is slower than ours in terms of latency (see Table~\ref{tab:MNIST-performance-comparison}).
In another recent work, \citet{badawi2018alexnet} obtained a $40.41\times$
acceleration over CryptoNets by using GPUs. In terms of latency, our
solution is more than $6\times$ faster than their solution even though
it uses only the CPU. \citet{sanyal2018tapas} argued that many of
these methods leak information about the structure of the neural-network
that the service provider uses through the parameters of the encryption.
They presented a method that leaks less information about the neural-network
but their solution is considerably slower. Nevertheless, their solution
has the nice benefit that it allows the service provider to change
the network without requiring changes on the client side.

In parallel to our work, \citet{chou2018faster} introduce alternative
methods to reduce latency.  Their solution on MNIST runs in 39.1s
(98.7\% accuracy), whereas our LoLa runs in 2.2s (98.95\% accuracy).
On CIFAR, their inference takes more than 6 \textit{hours} (76.7\%
accuracy), and ours takes less than 12 \textit{minutes} (74.1\% accuracy).
\citet{makri2019epic} apply transfer learning to private inference.
However, their methods and threat model are different.

Other researchers proposed using different encryption schemes. For
example, the Chameleon system \citep{riazi2018chameleon} uses MPC
to demonstrate private predictions on MNIST and \citet{juvekar2018gazelle}
used a hybrid MPC-HE approach for the same task. Several hardware
based solutions, such as the one of \citet{tramer2018slalom} were
proposed. These approaches are faster however, this comes at the cost
of lower level of security.

\section{Background}

In this section we provide a concise background on Homomorphic Encryption
and CryptoNets. We refer the reader to \citet{dowlin2017manual} and
\citet{gilad2016cryptonets} for a detailed exposition. We use bold
letters to denote vectors and for any vector $\mathbf{u}$ we denote
by $u_{i}$ its $i^{\text{th }}$coordinate.

\subsection{Homomorphic Encryption\label{subsec:Homomrphic-Encryption}}

In this study, we use Homomorphic Encryptions (HE) to provide privacy.
HE are encryptions that allow operations on encrypted data without
requiring access to the secret key \citep{Gen09}. The data used for
encryption is assumed to be elements in a ring $\ring$. On top of
the encryption function $\EE$ and the decryption function $\DD$,
the HE scheme provides two additional operators $\oplus$ and $\otimes$
so that for any $x_{1},x_{2}\in\ring$
\begin{align*}
\DD\left(\EE\text{\ensuremath{\left(x_{1}\right)}\ensuremath{\ensuremath{\oplus\EE\left(x_{2}\right)}}}\right) & =x_{1}+x_{2}\,\,\text{\,{and}}\\
\DD\left(\EE\left(x_{1}\right)\otimes\EE\left(x_{2}\right)\right) & =x_{1}\times x_{2}
\end{align*}
where $+$ and $\times$ are the standard addition and multiplication
operations in the ring $\ring$. Therefore, the $\oplus$ and $\otimes$
operators allow computing addition and multiplication on the data
in its encrypted form and thus computing any polynomial function.
We note that the ring $\ring$ refers to the plaintext message ring
and not the encrypted message ring. In the rest of the paper we refer
to the former ring and note that by the homomorphic properties, any
operation on an encrypted message translates to the same operation
on the corresponding plaintext message.

Much progress has been made since \citet{Gen09} introduced the first
HE scheme. We use the Brakerski/Fan-Vercauteren scheme\footnote{We use version 2.3.1 of the  SEAL, \url{http://sealcrypto.org/},
	with parameters that guarantee 128 bits of security according to the
	proposed standard for Homomorphic Encryptions \citep{albrecht2018homomorphic}.} (BFV) \citep{fan2012somewhat,brakerski2014efficient} In this scheme,
the ring on which the HE operates is $\ring=\frac{\Z_{p}[x]}{x^{n}+1}$
where $\Z_{p}=\frac{\Z}{p\Z}$, and $n$ is a power of $2$.\footnote{See supplementary material for a brief introduction to rings used
	in this work.} If the parameters $p$ and $n$ are chosen so that there is an order
$2n$ root of unity in $\Z_{p}$, then every element in $\ring$ can
be viewed as a vector of dimension $n$ of elements in $\Z_{p}$ where
addition and multiplication operate component-wise \citep{brakerski2014leveled}.
Therefore, throughout this essay we will refer to the messages as
$n$ dimensional vectors in $\Z_{p}$. The BFV scheme allows another
operation on the encrypted data: rotation. The rotation operation
is a slight modification to the standard rotation operation of size
$k$ that sends the value in the $i$'th coordinate of a vector to
the $\left(\left(i+k\right)\mod n\right)$ coordinate.

To present the rotation operations it is easier to think of the message
as a $2\times\nicefrac{n}{2}$ matrix:

\[
\begin{bmatrix}m_{1} & m_{2} & \cdot & \cdot & m_{\nicefrac{n}{2}}\\
m_{\nicefrac{n}{2}+1} & m_{\nicefrac{n}{2}+2} & \cdot & \cdot & m_{n}
\end{bmatrix}
\]
with this view in mind, there are two rotations allowed, one switches
the row, which will turn the above matrix to
\[
\begin{bmatrix}m_{\nicefrac{n}{2}+1} & m_{\nicefrac{n}{2}+2} & \cdot & \cdot & m_{n}\\
m_{1} & m_{2} & \cdot & \cdot & m_{\nicefrac{n}{2}}
\end{bmatrix}
\]
and the other rotates the columns. For example, rotating the original
matrix by one column to the right will result in
\[
\begin{bmatrix}m_{\nicefrac{n}{2}} & m_{1} & \cdot & \cdot & m_{\nicefrac{n}{2-1}}\\
m_{n} & m_{\nicefrac{n}{2}+1} & \cdot & \cdot & m_{n-1}
\end{bmatrix}\,\,\,.
\]

Since $n$ is a power of two, and the rotations we are interested
in are powers of two as well, thinking about the rotations as standard
rotations of the elements in the message yields similar results for
our purposes. In this view, the row-rotation is a rotation of size
$\nicefrac{n}{2}$ and smaller rotations are achieved by column rotations.

\subsection{CryptoNets\label{subsec:CryptoNets}}

\citet{gilad2016cryptonets} introduced CryptoNets, a method that
performs private predictions on neural networks using HE. Since HE
does not support non-linear operations such as ReLU or sigmoid, they
replaced these operations  by \textit{square} activations. Using a
convolutional network with 4 hidden layers, they demonstrated that
CryptoNets can make predictions with an accuracy of $99\%$ on the
MNIST task. They presented latency of $205$ seconds for a single
prediction and throughput of $\sim59000$ predictions per hour.

The high throughput is due to the vector structure of encrypted messages
used by CryptoNets, which allows processing multiple records simultaneously.
CryptoNets use a message for every input feature. However, since each
message can encode a vector of dimension $n$, then $n$ input records
are encoded simultaneously such that $v_{j}^{i}$, which is the value
of the $j$'th feature in the $i$'th record is mapped to the $i$'th
coordinate of the $j$'th message. In CryptoNets all operations between
vectors and matrices are implemented using additions and multiplications
only (no rotations). For example, a dot product between two vectors
of length $k$ is implemented by $k$ multiplications and additions. 

CryptoNets have several limitations. First, the fact that each feature
is represented as a message, results in a large number of operations
for a single prediction and therefore high latency. The large number
of messages also results in high memory usage and creates memory bottlenecks
as we demonstrate in Section~\ref{sec:Cifar}. Furthermore, CryptoNets
cannot be applied to deep networks such as AlexNet. This is because
the multiplication operation in each layer increases the noise in
the encrypted message and the required size of each message, which
makes each operation significantly slower when many layers exist in
the network (see Section \ref{sec:Applying-Deep-Nets} for further
details).

\section{LoLa}

In this section we present the Low-Latency CryptoNets (LoLa) solution
for private inference. LoLa uses various representations of the encrypted
messages and alternates between them during the computation. This
results in better latency and memory usage than CryptoNets, which
uses a single representation where each pixel (feature) is encoded
as a separate message.

In Section~\ref{subsec:Linear-Classifier-Example}, we show a simple
example of a linear classifier, where a change of representation can
substantially reduce latency and memory usage. In Section~\ref{subsec:Message-Representations},
we present different types of representations and how various matrix-vector
multiplication implementations can transform one type of representation
to another. In Section~\ref{sec:MNIST}, we apply LoLa in a private
inference task on MNIST and show that it can perform a single prediction
in 2.2 seconds. In Section~\ref{sec:Cifar}, we apply LoLa to perform
private inference on the CIFAR-10 dataset.

\subsection{Linear Classifier Example\label{subsec:Linear-Classifier-Example}}

In this section we provide a simple example of a linear classifier
that illustrates the limitations of CryptoNets that are due to its
representation. We show that a simple change of the input representation
results in much faster inference and significantly lower memory usage.
This change of representation technique is at the heart of the LoLa
solution and in the next sections we show how to extend it to non-linear
neural networks with convolutions.

Assume that we have a single $d$ dimensional input vector $\vv$
(e.g., a single MNIST image) and we would like to apply a private
prediction of a linear classifier $\ww$ on $\vv$. Using the CryptoNets
representation, we would need $d$ messages for each entry of $\vv$
and $d$ multiplication and addition operations to perform the dot
product $\ww\cdot\vv$.

Consider the following alternative representation: encode the input
vector $\vv$ as a single message $\mm$ where for all $i$, $m_{i}=v_{i}$.
We can implement a dot product between two vectors, whose sizes are
a power of $2$, by applying point-wise multiplication between the
two vectors and a series of $\log d$ rotations of size $1,2,4,\ldots,\nicefrac{d}{2}$
and addition between each rotation. The result of such a dot product
is a vector that holds the results of the dot-product in all its
coordinates.\footnote{Consider calculating the dot product of the vectors $\left(v_{1},...,v_{4}\right)$
	and $\left(w_{1},...,w_{4}\right).$ Point-wise multiplication, rotation
	of size 1 and summation results in the vector $\left(v_{1}w_{1}+v_{4}w_{4},v_{2}w_{2}+v_{1}w_{1},v_{3}w_{3}+v_{2}w_{2},v_{4}w_{4}+v_{3}w_{3}\right)$.
	Another rotation of size 2 and summation results in the 4 dimensional
	vector which contains the dot-product of the vectors in all coordinates.} Thus, in total the operation requires $\log d$ rotations and additions
and a single multiplication, and uses only a single message. This
results in a significant reduction in latency and memory usage compared
to the CryptoNets approach which requires $d$ operations and $d$
messages.

\subsection{Message Representations\label{subsec:Message-Representations}}

In the previous section we saw an example of a linear classifier with
two different representations of the encrypted data and how they affect
the number of HE operations and memory usage. To extend these observations
to generic feed-forward neural networks, we define other forms of
representations which are used by LoLa. Furthermore, we define different
implementations of matrix-vector multiplications and show how they
change representations during the forward pass of the network. As
an example of this change of representation, consider a matrix-vector
multiplication, where the input vector is represented as a single
message ($m_{i}=v_{i}$ as in the previous section). We can multiply
the matrix by the vector using $r$ dot-product operations, where
the dot-product is implemented as in the previous section, and $r$
is the number of rows in the matrix. The result of this operation
is a vector of length $r$ that is spread across $r$ messages (recall
that the result of the dot-product operation is a vector which contains
the dot-product result in all of its coordinates). Therefore, the
result has a different representation than the representation of the
input vector. 

Different representations can induce different computational costs
and therefore the choice of the representations used throughout the
computation is important for obtaining an efficient solution. In the
LoLa solution, we propose using different representations for different
layers of the network. In this study we present implementations for
several neural networks to demonstrate the gains of using varying
representations. Automating the process of finding efficient representations
for a given network is beyond the scope of the current work.

We present different possible vector representations in Section~\ref{subsec:Vector-representations}
and discuss matrix-vector multiplication implementations in Section~\ref{subsec:Matrix-vector-multiplications}.
We note that several representations are new (e.g., convolutional
and interleaved), whereas SIMD and dense representations were used
before.

\subsubsection{Vector representations\label{subsec:Vector-representations}}

Recall that a message  encodes a vector of length $n$ of elements
in $\Z_{p}$. For the sake of brevity, we assume that the dimension
of the vector $\vv$ to be encoded is of length $k$ such that $k\leq n$,
for otherwise multiple messages can be combined. We will consider
the following representations:

\paragraph{Dense representation.}

A vector $\vv$ is represented as a single message $\mm$ by setting
$v_{i}\mapsto m_{i}$.

\paragraph{Sparse representation.}

A vector $\vv$ of length $k$ is represented in $k$ messages $\mm^{1},\ldots\mm^{k}$
such that $\mm^{i}$ is a vector in which every coordinate is set
to $v_{i}$.\footnote{Recall that the messages are in the ring $R=\frac{\Z_{p}[x]}{x^{n}+1}$
	which, by the choice of parameters, is homomorphic to $\left(\Z_{p}\right)^{n}$.
	When a vector has the same value $v_{i}$ in all its coordinates,
	then its polynomial representation in $\frac{\Z_{p}[x]}{x^{n}+1}$
	is the constant polynomial $v_{i}$.} 

\paragraph{Stacked representation.}

For a short (low dimension) vector $\vv$, the stacked representation
holds several copies of the vector $\vv$ in a single message $\mm$.
Typically this will be done by finding $d=\left\lceil \log\left(k\right)\right\rceil $,
the smallest $d$ such that the dimension of $\vv$ is at most $2^{d}$
and setting $m_{i},m_{i+2^{d}},m_{i+2\cdot2^{d}},\ldots=v_{i}.$

\paragraph{Interleaved representation.}

The interleaved representation uses a permutation $\sigma$ of $\left[1,\ldots,n\right]$
to set $m_{\sigma\left(i\right)}=v_{i}$. The dense representation
is a special case of the interleaved representation where $\sigma$
is the identity permutation.

\paragraph{Convolution representation. }

This is a special representation that makes convolution operations
efficient. A convolution, when flattened to a single dimension, can
be viewed as a restricted linear operation where there is a weight
vector $\ww$ of length $r$ (the window size) and a set of permutations
$\sigma_{i}$ such that the $i$\textquoteright th output of the linear
transformation is $\sum_{j}w_{j}v_{\sigma_{i}(j)}$. The convolution
representation takes a vector $v$ and represents it as $r$ messages
$\mm^{1},\ldots,\mm^{r}$ such that $m_{i}^{j}=v_{\sigma_{i}\left(j\right)}.$\footnote{For example, consider a matrix $A\in\mathbb{R}^{4\times4}$ which
	corresponds to an input image and a $2\times2$ convolution filter
	that slides across the image with stride 2 in each direction. Let
	$a_{i,j}$ be the entry at row $i$ and column $j$ of the matrix
	$A$. Then, in this case $r=4$ and the following messages are formed
	$M^{1}=(a_{1,1},a_{1,3},a_{3,1},a_{3,3})$ , $M^{2}=(a_{1,2},a_{1,4},a_{3,2},a_{3,4})$,
	$M^{3}=(a_{2,1},a_{2,3},a_{4,1},a_{4,3})$ and $M^{4}=(a_{2,2},a_{2,4},a_{4,2},a_{4,4})$.
	In some cases it will be more convenient to combine the interleaved
	representation with the convolution representation by a permutation
	$\tau$ such that $m_{\tau\left(i\right)}^{j}=v_{\sigma_{i}\left(j\right)}$.\label{fn:conv}}

\paragraph{SIMD representation:}

This is the representation used by CryptoNets \citep{gilad2016cryptonets}.
They represent each feature as a separate message but map multiple
data vectors into the same set of messages, as described in Section~\ref{subsec:CryptoNets}.

\subsubsection{Matrix-vector multiplications\label{subsec:Matrix-vector-multiplications}}

Matrix-vector multiplication is a core operation in neural networks.
The matrix may contain the learned weights of the network and the
vector represents the values of the nodes at a certain layer. Here
we present different ways to implement such matrix-vector operations.
Each method operates on vectors in different representations and produces
output in yet another representation. Furthermore, the weight matrix
has to be represented appropriately as a set of vectors, either column-major
or row-major to allow the operation. We assume that the matrix $W$
has $k$ columns $\cc^{1},\ldots,\cc^{k}$ and $r$ rows $\rr^{1},\ldots,\rr^{r}.$
We consider the following matrix-vector multiplication implementations.

\paragraph{Dense Vector -- Row Major.}

If the vector is given as a dense vector and each row $\rr^{j}$ of
the weight matrix is encoded as a dense vector then the matrix-vector
multiplication can be applied using $r$ dot-product operations. As
already described above, a dot-product requires a single multiplication
and $\log\left(n\right)$ additions and rotations. The result is a
sparse representation of a vector of length $r$.

\paragraph{Sparse Vector -- Column Major.}

Recall that $W\vv=\sum v_{i}\cc^{i}$. Therefore, when $\vv$ is encoded
in a sparse format, the message $\mm^{i}$ has all its coordinate
set to $v_{i}$ and $v_{i}\cc^{i}$ can be computed using a single
point-wise multiplication. Therefore, $W\vv$ can be computed using
$k$ multiplications and additions and the result is a dense vector.

\paragraph{Stacked Vector -- Row Major.}

For the sake of clarity, assume that $k=2^{d}$ for some $d$. In
this case $\nicefrac{n}{k}$ copies of $\vv$ can be stacked in a
single message $\mm$ (this operation requires $\log\left(\nicefrac{n}{k}\right)-1$
rotations and additions). By concatenating $\nicefrac{n}{k}$ rows
of $W$ into a single message, a special version of the dot-product
operation can be used to compute $\nicefrac{n}{k}$ elements of $W\vv$
at once. First, a point-wise multiplication of the stacked vector
and the concatenated rows is applied followed by $d-1$ rotations
and additions where the rotations are of size $1,2,\ldots,2^{d-1}$.
The result is in the interleaved representation.\footnote{For example, consider a $2\times2$ matrix $W$ flattened to a vector
	$\mathbf{w}=(w_{1,1},w_{1,2},w_{2,1},w_{2,2})$ and a two-dimensional
	vector $\mathbf{v}=(v_{1},v_{2})$. Then, after stacking the vectors,
	point-wise multiplication, rotation of size $1$ and summation, the
	second entry of the result contains $w_{1,1}v_{1}+w_{1,2}v_{2}$ and
	the fourth entry contains $w_{2,1}v_{1}+w_{2,2}v_{2}$. Hence, the
	result is in an interleaved representation.} 

The Stacked Vector - Row Major gets its efficiency from two places.
First, the number of modified dot product operations is $\nicefrac{rk}{n}$
and second, each dot product operation requires a single multiplication
and only $d$ rotations and additions (compared to $\log n$ rotations
and additions in the standard dot-product procedure).

\paragraph{Interleaved Vector -- Row Major.}

This setting is very similar to the dense vector -- row major matrix
multiplication procedure with the only difference being that the columns
of the matrix have to be shuffled to match the permutation of the
interleaved representation of the vector. The result is in sparse
format.

\paragraph{Convolution vector -- Row Major.\label{subsec:Convolution-vector-=002013row major}}

A convolution layer applies the same linear transformation to different
locations on the data vector $\vv$. For the sake of brevity, assume
the transformation is one-dimensional. In neural network language
that would mean that the kernel has a single map. Obviously, if more
maps exist, then the process described here can be repeated multiple
times. 

Recall that a convolution, when flattened to a single dimension, is
a restricted linear operation where the weight vector $\ww$ is of
length $r$, and there exists a set of permutations $\sigma_{i}$
such that the $i$\textquoteright th output of the linear transformation
is $\sum w_{j}v_{\sigma_{i}(j)}$. In this case, the convolution representation
is made of $r$ messages such that the $i$\textquoteright th element
in the message $\mm^{j}$ is $v_{\sigma_{i}(j)}$. By using a sparse
representation of the vector $\ww$, we get that $\sum w_{j}\mm^{j}$
computes the set of required outputs using $r$ multiplications and
additions. When the weights are not encrypted, the multiplications
used here are relatively cheap since the weights are scalar and BFV
supports fast implementation of multiplying a message by a scalar.
The result of this operation is in a dense format.

\subsection{Secure Networks for MNIST\label{sec:MNIST}}

Here we present private predictions on the MNIST data-set \citep{lecun2010mnist}
using the techniques described above and compare it to other private
prediction solutions for this task (see Table~\ref{tab:MNIST-performance-comparison}).
Recall that CryptoNets use the SIMD representation in which each pixel
requires its own message. Therefore, since each image in the MNIST
data-set is made up of an array of $28\times28$ pixels, the input
to the CryptoNets network is made of $784$ messages. On the reference
machine used for this work (Azure standard B8ms virtual machine with
8 vCPUs and 32GB of RAM), the original CryptoNets implementation runs
in 205 seconds. Re-implementing it to use better memory management
and multi-threading in SEAL 2.3 reduces the running time to $24.8$
seconds. We refer to the latter version as CryptoNets~2.3.

\begin{table*}
	\begin{centering}
		\caption{Message size, message representation and operations in each layer
			of the LoLa inference solution on MNIST. The input size format is
			number of vectors $\times$ dimension. \label{tab:LoLa-data-representation-1}}
		\par\end{centering}
	\vskip 0.15in
	\begin{centering}
		\begin{tabular}{|c|c|c|c|}
			\hline 
			\textbf{Layer} & \textbf{Input size } & \textbf{Representation} & \textbf{LoLa operation}\tabularnewline
			\hline 
			\hline 
			\multirow{2}{*}{$5\times5$ convolution layer} & $25\times169$ & convolution & convolution vector -- row major multiplication\tabularnewline
			\cline{2-4} 
			& $5\times169$ & dense & combine to one vector using $4$ rotations and additions\tabularnewline
			\hline 
			square layer & $1\times845$ & dense & square\tabularnewline
			\hline 
			\multirow{3}{*}{dense layer} & $1\times845$ & dense & stack vectors using $8$ rotations and additions\tabularnewline
			\cline{2-4} 
			& $1\times6760$ & stacked & stacked vector -- row major multiplication\tabularnewline
			\cline{2-4} 
			& $13\times8$ & interleave & combine to one vector using 12 rotations and additions\tabularnewline
			\hline 
			square layer & $1\times100$ & interleave & square\tabularnewline
			\hline 
			dense layer & $1\times100$ & interleave & interleaved vector -- row major\tabularnewline
			\hline 
			output layer & $1\times10$ & sparse & \tabularnewline
			\hline 
		\end{tabular}
		\par\end{centering}
	\vskip -0.1in
\end{table*}

LoLa and CryptoNets use different approaches to evaluating neural
networks. As a benchmark, we applied both to the same network that
has accuracy of $98.95\%$. After suppressing adjacent linear layers
it can be presented as a $5\times5$ convolution layer with a stride
of $(2,2)$ and $5$ output maps, which is followed by a square activation
function that feeds a fully connected layer with $100$ output neurons,
another square activation and another fully connected layer with $10$
outputs (in the supplementary material we include an image of the
architecture). 

LoLa uses different representations and matrix-vector multiplication
implementations throughout the computation. Table~\ref{tab:LoLa-data-representation-1}
summarizes the message representations and operations that LoLa applies
in each layer. The inputs to LoLa are $25$ messages which are the
convolution representation of the image. Then, LoLa performs a convolution
vector -- row major multiplication for each of the $5$ maps of the
convolution layer which results in $5$ dense output messages. These
$5$ dense output messages are joined together to form a single dense
vector of $5*169=845$ elements. This vector is squared using a single
multiplication and $8$ copies of the results are stacked before applying
the dense layer. Then $13$ rounds of Stacked vector -- Row Major
multiplication are performed. The $13$ vectors of interleaved results
are rotated and added to form a single interleaved vector of dimension
$100$. The vector is then squared using a single multiplication.
Finally, Interleaved vector -- Row Major multiplication is used to
obtain the final result in sparse format.

LoLa computes the entire network in only $2.2$ seconds which is $11\times$
faster than CryptoNets~2.3 and $93\times$ faster than CryptoNets.
Table~\ref{tab:MNIST-performance-comparison} shows a summary of
the performance of different methods. In the supplementary material
we show the dependence of the performance on the number of processor
cores. We provide two additional versions of LoLa. The first, LoLa-Dense,
uses a dense representation as input and then transforms it to a convolutional
representation using HE operations. Then it proceeds similarly to
LoLa in subsequent layers. It performs a single prediction in 7.2
seconds. We provide more details on this solution in the supplementary
material. The second version, LoLa-Small is similar to Lola-Conv but
has only a convolution layer, square activation and a dense layer.
This solution has an accuracy of only $96.92\%$ but can make a prediction
in as little as $0.29$ seconds.

\begin{table*}
	\caption{MNIST performance comparison. Solutions are grouped by accuracy levels.
		\label{tab:MNIST-performance-comparison}}
	
	\vskip 0.15in
	\begin{centering}
		\begin{tabular}{cccc}
			\toprule 
			\textbf{Method} & \textbf{Accuracy} & \textbf{Latency} & \tabularnewline
			\midrule
			\midrule 
			FHE--DiNN100 & 96.35\% & 1.65 & \citep{bourse2017fast}\tabularnewline
			\midrule 
			LoLa-Small & 96.92\% & \textbf{0.29} & \tabularnewline
			\midrule
			\midrule 
			CryptoNets & 98.95\% & 205 & \citep{gilad2016cryptonets}\tabularnewline
			\midrule 
			nGraph-HE & $98.95\%$\footnotemark & 135 & \citep{boemer2018ngraph}\tabularnewline
			\midrule 
			Faster-CryptoNets & 98.7\% & 39.1 & \citep{chou2018faster}\tabularnewline
			\midrule 
			CryptoNets 2.3 & 98.95 & 24.8 & \tabularnewline
			\midrule 
			HCNN & 99\% & 14.1 &  \citep{badawi2018alexnet}\tabularnewline
			\midrule 
			LoLa-Dense & 98.95\% & 7.2 & \tabularnewline
			\midrule 
			LoLa & 98.95\% & \textbf{2.2} & \tabularnewline
			\bottomrule
		\end{tabular}
		\par\end{centering}
	\vskip -0.1in
\end{table*}

\footnotetext{The accuracy is not reported in \citet{boemer2018ngraph}. However, they implement the same network as in \citet{gilad2016cryptonets}.}

\subsection{Secure Networks for CIFAR\label{sec:Cifar}}

The Cifar-10 data-set \citep{krizhevsky2009learning} presents a more
challenging task of recognizing one of 10 different types of objects
in an image of size $3\times32\times32$ . For this task we train
a convolutional neural network that is depicted in Figure~\ref{fig:cifar-full}.
The exact details of the architecture are given in the supplementary
material. 
\begin{figure*}
	\includegraphics[viewport=-50bp 0bp 650bp 400bp,width=0.8\textwidth,height=0.3\textheight]{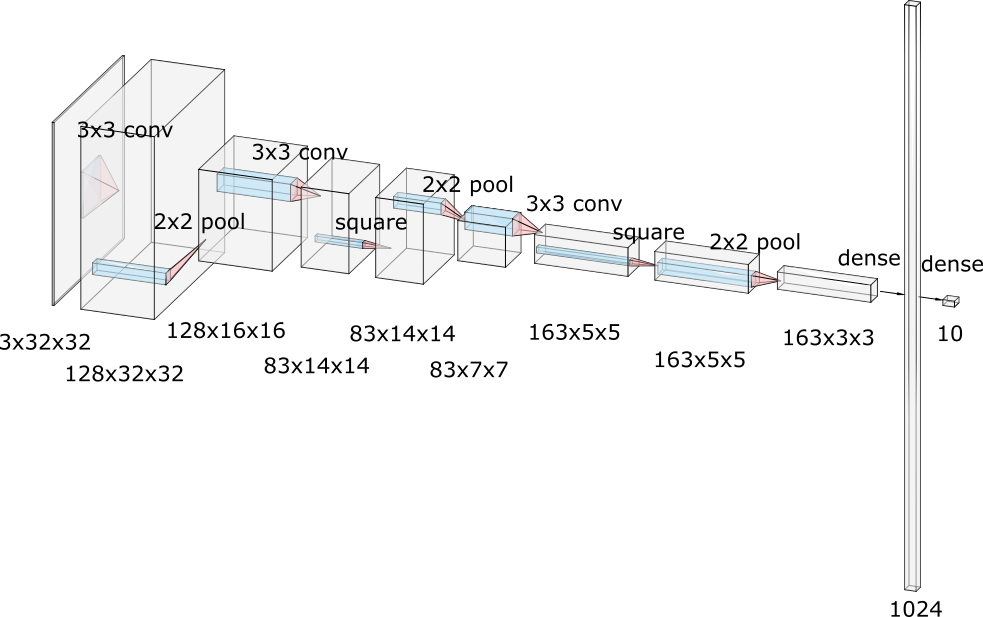}\caption{The structure of the network used for CIFAR classification.\label{fig:cifar-full}}
\end{figure*}
For inference, adjacent linear layers were collapsed to form a network
with the following structure: (i) $8\times8\times3$ convolutions
with a stride of $\left(2,2\right)$ and $83$ maps (ii) square activation
(iii) $6\times6\times83$ convolution with stride $\left(2,2\right)$
and $163$ maps (iv) square activation (v) dense layer with $10$
output maps. The accuracy of this network is $74.1\%$. This network
is much larger than the network used for MNIST by CryptoNets. The
input to this network has $3072$ nodes, the first hidden layer has
$16268$ nodes and the second hidden layer has $4075$ nodes (compared
to $784,\,845$, and $100$ nodes respectively for MNIST). Due to
the sizes of the hidden layers, implementing this network with the
CryptoNet approach of using the SIMD representation requires 100's
GB of RAM since a message has to be memorized for every node. Therefore,
this approach is infeasible on most computers.

For this task we take a similar approach to the one presented in Section~\ref{sec:MNIST}.
The image is encoded using the convolution representation into $3\times8\times8=192$
messages. The convolution layer is implemented using the convolution
vector -- row major matrix-vector multiplication technique. The results
are combined into a single message using rotations and additions which
allows the square activation to be performed with a single point-wise
multiplication. The second convolution layer is performed using row
major-dense vector multiplication. Although this layer is a convolution
layer, each window of the convolution is so large that it is more
efficient to implement it as a dense layer. The output is a sparse
vector which is converted into a dense vector by point-wise multiplications
and additions which allows the second square activation to be performed
with a single point-wise multiplication. The last dense layer is implemented
with a row major-dense vector technique again resulting in a sparse
output.

LoLa uses plain-text modulus $p=2148728833\times2148794369\times2149810177$
(the factors are combined using the Chinese Reminder Theorem) and
$n=16384$. During the computation, LoLa uses 12 GB of RAM for a single
prediction. It performs a single prediction in $730$ seconds out
of which the second layer consumes $711$ seconds. The bottleneck
in performance is due to the sizes of the weight matrices and data
vectors as evident by the number of parameters which is $>500,000$,
compared to $<90,000$ in the MNIST network.

\section{Private Inference using Deep Representations\label{sec:Applying-Deep-Nets}}

Homomorphic Encryptions have two main limitations when used for evaluating
deep networks: noise growth and message size growth. Every encrypted
message contains some noise and every operation on encrypted message
increases the noise level. When the noise becomes too large, it is
no longer possible to decrypt the message correctly. The mechanism
of \emph{bootstrapping} \citep{Gen09} can mitigate this problem but
at a cost of a performance hit. The message size grows with the size
of the network as well. Since, in its core, the HE scheme operates
in $\Z_{p}$, the parameter $p$ has to be selected such that the
largest number obtained during computation would be smaller than $p$.
Since every multiplication might double the required size of $p$,
it has to grow exponentially with respect to the number of layers
in the network. The recently introduced HEAAN scheme \citep{cheon2017homomorphic}
is more tolerant towards message growth but even HEAAN would not be
able to operate efficiently on deep networks.

We propose solving both the message size growth and the noise growth
problems using deep representations: Instead of encrypting the data
in its raw format, it is first converted, by a standard network, to
create a deep representation. For example, if the data is an image,
then instead of encrypting the image as an array of pixels, a network,
such as AlexNet \citep{krizhevsky2012imagenet}, VGG \citep{simonyan2014very},
or ResNet \citep{he2016deep}, first extracts a deep representation
of the image, using one of its last layers. The resulting representation
is encrypted and sent for evaluation. This approach has several advantages.
First, this representation is small even if the original image is
large. In addition, with deep representations it is possible to obtain
high accuracies using shallow networks: in most cases a linear predictor
is sufficient which translates to a fast evaluation with HE. It is
also a very natural thing to do since in many cases of interest, such
as in medical image, training a very deep network from scratch is
almost impossible since data is scarce. Hence, it is a common practice
to use deep representations and train only the top layer(s) \citep{yosinski2014transferable,tajbakhsh2016convolutional}. 

To test the deep representation approach we used AlexNet \citep{krizhevsky2012imagenet}
to generate features and trained a linear model to make predictions
on the CalTech-101 data-set \citep{calech101}.\footnote{More complex classifiers did not improve accuracy.}
In the supplementary material we provide a summary of the data representations
used for the CalTech-101 dataset. Since the CalTech-101 dataset is
not class balanced, we used only the first 30 images from each class
where the first $20$ where used for training and the other $10$
examples where used for testing. The obtained model has class-balanced
accuracy of $81.6\%$. The inference time, on the encrypted data,
takes only $0.16$ seconds when using the dense vector -- row major
multiplication. We note that such transfer learning approaches are
common in machine learning but to the best of our knowledge were not
introduced as a solution to private predictions with He before.

The use of transfer learning for private predictions has its limitations.
For example, if a power-limited client uses private predictions to
offload computation to the cloud, the transfer learning technique
would not be useful because most of the computation is on the client's
side. However, there are important cases in which this technique is
useful. For example, consider a medical institution which trains a
shallow network on deep representations of private x-ray images of
patients, and would like to make its model available for private predictions.
However, to protect its intellectual property, it is not willing to
share its model. In that case, it can use this technique to provide
private predictions while protecting the privacy of the model.

\section{Conclusions }

The problem of privacy in machine learning is gaining importance due
to legal requirements and greater awareness to the benefits and risks
of machine learning systems. In this study, we presented two HE based
solutions for private inference that address key limitations of previous
HE based solutions. We demonstrated both the ability to operate on
more complex networks as well as lower latency on networks that were
already studied in the past. 

The performance gain is mainly due to the use of multiple representations
during the computation process. This may be useful in other applications
of HE. One example is training machine learning models over encrypted
data. This direction is left for future study.

\section*{Acknowledgments}

We thank Kim Laine for helpful discussions.

\bibliographystyle{iclr2019_conference}
\bibliography{LoLa}

\appendix

\section{Rings}

In this work we consider commutative rings $\ring$. A ring is a set
which is equipped with addition and multiplication operations and
satisfies several ring axioms such as $a+b=b+a$ for all $a,b\in\ring$.
A commutative ring is a ring in which the multiplication is commutative.,
i.e., $ab=ba$ for all $a,b\in\ring$. Since all rings we consider
are commutative, we use the term ``ring'' to refer to a commutative
ring.

The set $\mathbb{Z}$ and the set $\mathbb{Z}_{p}$ of integers modulu
$p$ are rings. The elements of $\mathbb{Z}_{p}$ can be thought of
as sets of the form $\{i+ap:a\in\mathbb{Z}\}.$ The notation $k\in\mathbb{Z}_{p}$
refers to the set $\{k+ap:a\in\mathbb{Z}\}$. Alternatively, $k$
is a representative of the the set $\{k+ap:a\in\mathbb{Z}\}$. 

The set $\ring[x]$ of polynomials with coefficients in a ring $\ring$
is itself a ring. Thus, $\mathbb{Z}_{p}[x]$ is the ring of polynomials
with coefficients in $\mathbb{Z}_{p}[x]$. Finally, we introduce the
ring $\frac{\mathbb{Z}_{p}[x]}{x^{n}+1}$ which is the ring used in
the BFV scheme. The elements of this ring can be thought of as sets
of the form $\left\{ r(x)+q(x)\left(x^{n}+1\right)\mid q(x)\in\mathbb{Z}_{p}[x]\right\} $. 

The notation $t(x)\in\frac{\mathbb{Z}_{p}[x]}{x^{n}+1}$ refers to
the set $\left\{ t(x)+q(x)\left(x^{n}+1\right)\mid q(x)\in\mathbb{Z}_{p}[x]\right\} $.
Conversely, we say that $t(x)$ is a representative of the set $\left\{ t(x)+q(x)\left(x^{n}+1\right)\mid q(x)\in\mathbb{Z}_{p}[x]\right\} $.
For each element in $\frac{\mathbb{Z}_{p}[x]}{x^{n}+1}$, there is
a representative in $\mathbb{Z}_{p}[x]$ with degree at most $n-1$.
Furthermore, any two non-equal polynomials of degree at most $n-1$
in $\mathbb{Z}_{p}[x]$, are representatives of different elements
in $\frac{\mathbb{Z}_{p}[x]}{x^{n}+1}$.

\section{Parallel Scaling\label{sec:Parallel-Scaling}}

\begin{figure}
	\begin{centering}
		\includegraphics[height=5cm]{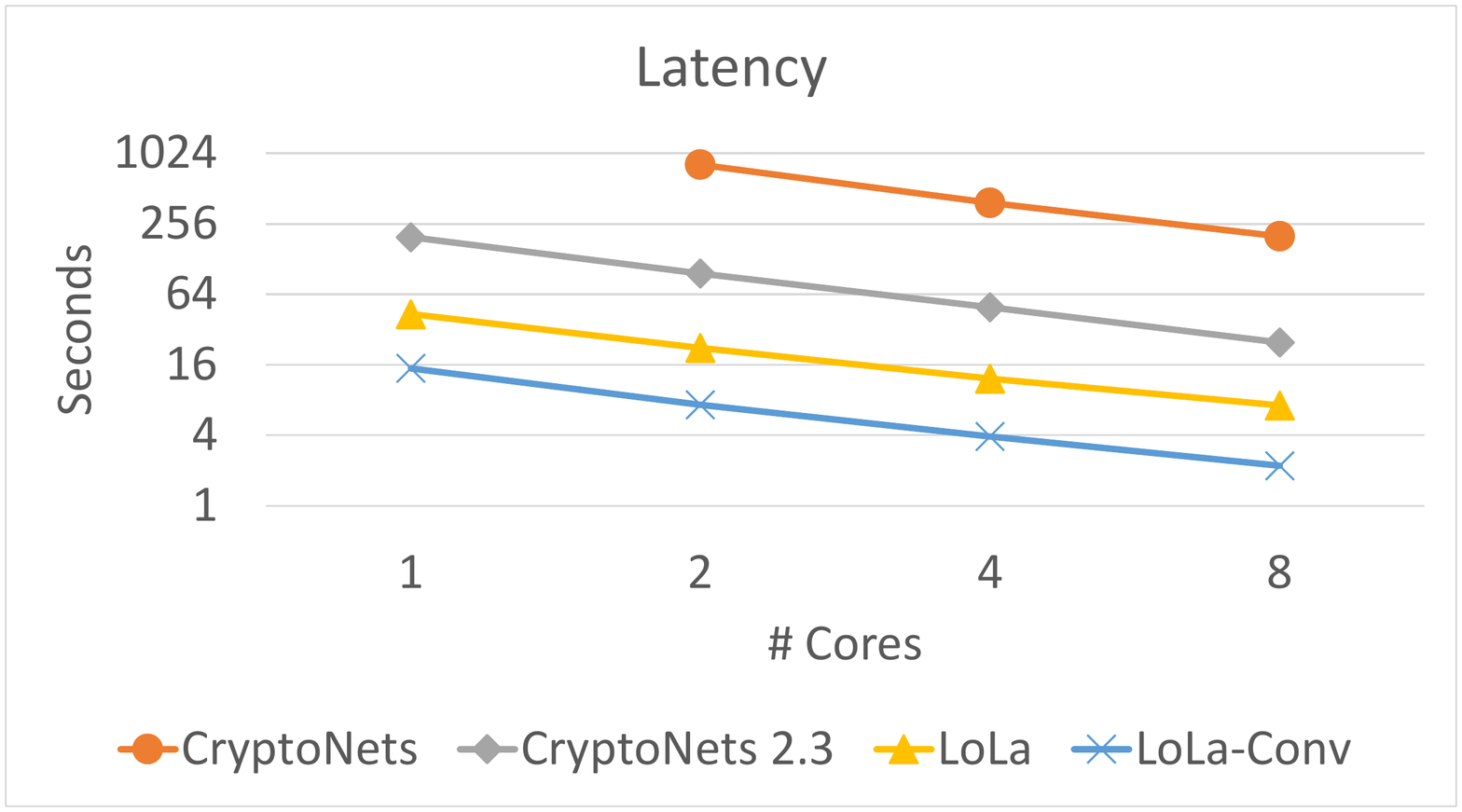}~~~\includegraphics[height=5cm]{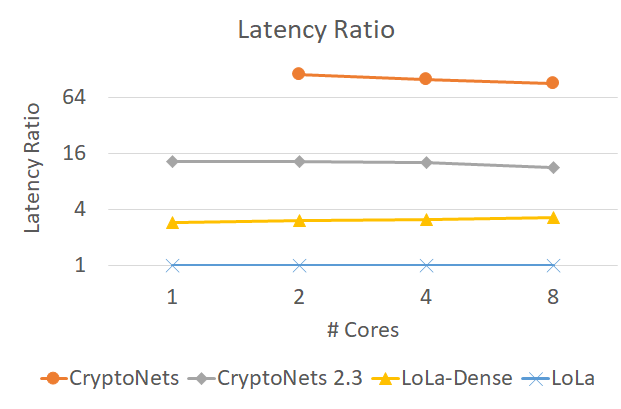}
		\par\end{centering}
	\caption{Left: latency of the different network implementations for the MNIST
		task with respect to the number of available cores. Right: ratio between
		the latency of each solution and the latency of LoLa \label{Figure: Scaling}}
\end{figure}

The performance of the different solutions is affected by the amount
of parallelism allowed. The hardware used for experimentation in this
work has 8 cores. Therefore, we tested the performance of the different
solutions with 1, 2, 4, and 8 cores to see how the performance varies.
The results of these experiments are presented in Figure~\ref{Figure: Scaling}.
These results show that at least up to $8$ cores the performance
of all methods scales linearly when tested on the MNIST data-set.
This suggests that the latency can be further improved by using machines
with higher core count. We note that the algorithm utilizes the cores
well and therefore we do not expect large gains from running multiple
queries simultaneously.

\section{Lola-Dense}

\begin{figure}
	\begin{center}
	\includegraphics[width=0.5\textwidth]{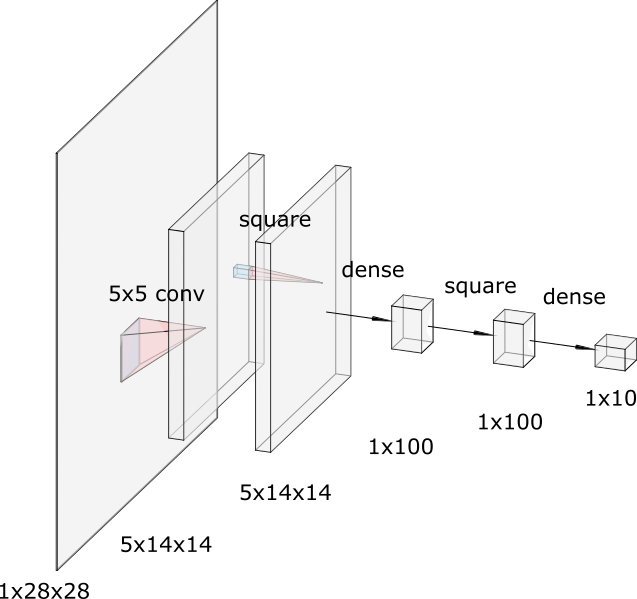}
	\end{center}
	\caption{The structure of the network used for MNIST classification.\label{fig:mnist}}
\end{figure}

LoLa-Dense uses the same network layout as CryptoNets (see Figure~\ref{fig:mnist})
and has accuracy of $98.95\%$. However, it is implemented differently:
the input to the network is a single dense message where the pixel
values are mapped to coordinates in the encoded vector line after
line . The first step in processing this message is breaking it into
25 messages corresponding to the 25 pixels in the convolution map
to generate a convolution representation. Creating each message requires
a single vector multiplication. This is performed by creating 25 masks.
The first mask is a vector of zeros and ones that corresponds to a
matrix of size $28\times28$ such that a one is in the $\left(i,j\right)$
coordinate if the $i,j$ pixel in the image appears as the upper left
corner of the $5\times5$ window of the convolution layer. Multiplying
point-wise the input vector by the mask creates the first message
in the convolution representation hybrided with the interleaved representation.
Similarly the other messages in the convolution representation are
created. Note that all masks are shifts of each other which allows
using the convolution representation-row major multiplication to implement
the convolution layer. To do that, think of the $25$ messages as
a matrix and the weights of a map of the convolution layer as a sparse
vector. Therefore, the outputs of the entire map can be computed using
$25$ multiplications (of each weight by the corresponding vector)
and $24$ additions. Note that there are $169$ windows and all of
them are computed simultaneously. However, the process repeats 5 times
for the $5$ maps of the convolution layer.

The result of the convolution layer are $5$ messages, each one of
them contains $169$ results. They are united into a single vector
by rotating the messages such that they will not have active values
in the same locations and summing the results. At this point, a single
message holds all the $845$ values ($169$ windows $\times5$ maps).
This vector is squared, using a single multiplication operation, to
implement the activation function that follows the convolution layer.
This demonstrates one of the main differences between CryptoNets and
LoLa; In CryptoNets, the activation layer requires $845$ multiplication
operations, whereas in LoLa it is a single multiplication. Even if
we add the manipulation of the vector to place all values in a single
message, as described above, we add only $4$ rotations and $4$ additions
which are still much fewer operations than in CryptoNets. 

Next, we apply a dense layer with $100$ maps. LoLa-Dense uses messages
of size $n=16384$ where the $845$ results of the previous layer,
even though they are in interleaving representation, take fewer than
$1024$ dimensions. Therefore, $16$ copies are stacked together which
allows the use of the Stacked vector -- Row Major multiplication
method. This allows computing $16$ out of the $100$ maps in each
operation and therefore, the entire dense layer is computed in 7 iterations
resulting in $7$ interleaved messages. By shifting the $i^{\text{th}}$
message by $i-1$ positions, the active outputs in each of the messages
are no longer in the same position and they are added together to
form a single interleaved message that contains the $100$ outputs.
The following square activation requires a single point-wise-multiplication
of this message. The final dense layer is applied using the Interleaved
vector -- Row Major method to generate $10$ messages, each of which
contains one of the $10$ outputs.\footnote{ It is possible, if required, to combine them into a single message
	in order to save communication. }

Overall, applying the entire network takes only $7.2$ seconds on
the same reference hardware which is $34.7\times$ faster than CryptoNets
and $3.4\times$ faster than CryptoNets 2.3.

\begin{table}
	\centering{}%
	\begin{tabular}{|c|c|c|c|}
		\hline 
		\textbf{Layer} & \textbf{Input size} & \textbf{Representation} & \textbf{LoLa-Dense Operation}\tabularnewline
		\hline 
		\hline 
		\multirow{3}{*}{$5\times5$ convolution layer} & $1\times784$ & dense & mask input to create 25 messages\tabularnewline
		\cline{2-4} 
		& $25\times169$ & convolution-interleave & convolution vector -- row major mult'\tabularnewline
		\cline{2-4} 
		& $5\times169$ & interleave & combine 5 messages into one\tabularnewline
		\hline 
		square layer & $1\times845$ & interleave & square\tabularnewline
		\hline 
		\multirow{3}{*}{dense layer} & $1\times845$ & interleave & stack 16 copies\tabularnewline
		\cline{2-4} 
		& $1\times13520$ & stacked-interleave & stacked vector -- row major mult'\tabularnewline
		\cline{2-4} 
		& $7\times16$ & interleave & combine 7 messages into one\tabularnewline
		\hline 
		square layer & $1\times100$ & interleave & square\tabularnewline
		\hline 
		dense layer & $1\times100$ & interleave & interleaved vector -- row major\tabularnewline
		\hline 
		output layer & $10\times1$ & sparse & \tabularnewline
		\hline 
	\end{tabular}\caption{Message size, message representation and operations in each layer
		of the LoLa-Dense inference solution on MNIST. The input size format
		is number of vectors $\times$ dimension\label{tab:LoLa-dense-data-representation}}
\end{table}

\section{Secure CIFAR}

The neural network used has the following layout: the input is a $3\times32\times32$
image (i) $3\times3$ linear convolution with stride of $\left(1,1\right)$
and 128 output maps, (ii) $2\times2$ average pooling with $\left(2,2\right)$
stride (iii) $3\times3$ convolution with $\left(1,1\right)$ stride
and 83 maps (iv) Square activation (v) $2\times2$ average pooling
with $\left(2,2\right)$ stride (vi) $3\times3$ convolution with
$\left(1,1\right)$ stride and 163 maps (vii) Square activation (vii)
$2\times2$ average pooling with stride $\left(2,2\right)$ (viii)
fully connected layer with $1024$ outputs (ix) fully connected layer
with $10$ outputs (x) softmax. ADAM was used for optimization \cite{kingma2014adam}
together with dropouts after layers (vii) and (viii). We use zero-padding
in layers (i) and (vii). See Figure~\ref{fig:cifar-full} for an
illustration of the network.

\begin{figure}
	\includegraphics[width=1\textwidth]{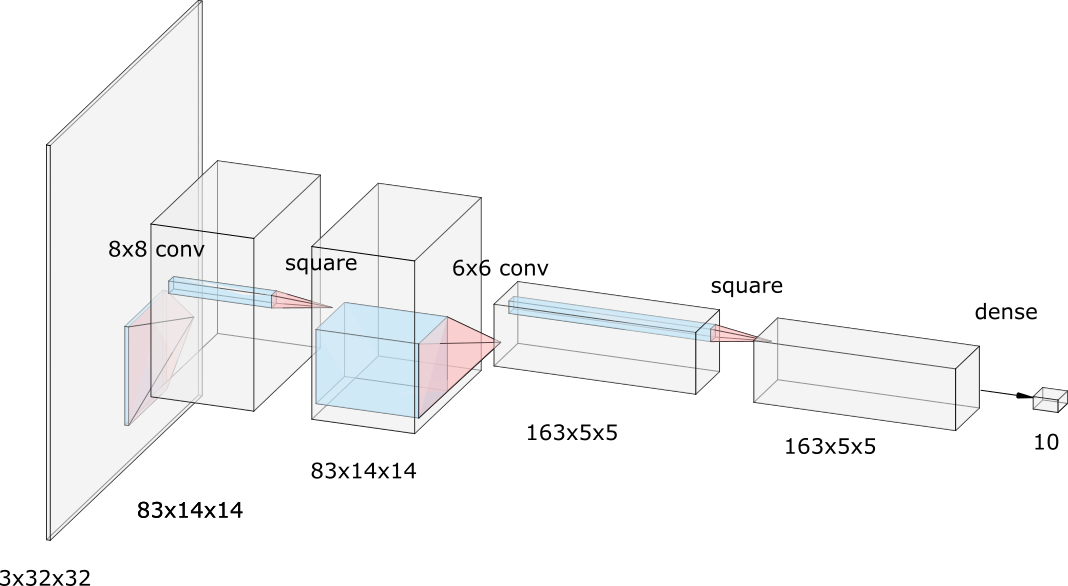}
	
	\caption{The structure of the network used for CIFAR classification after collapsing
		adjacent layers.\label{fig:cifar-collapsed}}
\end{figure}

For inference, adjacent linear layers were collapsed to form the following
structure: (i) $8\times8\times3$ convolutions with a stride of $\left(2,2,0\right)$
and 83 maps (ii) square activation (iii) $6\times6\times83$ convolution
with stride $\left(2,2,0\right)$ and 163 maps (iv) square activation
(v) dense layer with $10$ output maps. See Figure~\ref{fig:cifar-collapsed}
for an illustration.

\begin{figure}
	\includegraphics[width=1\textwidth]{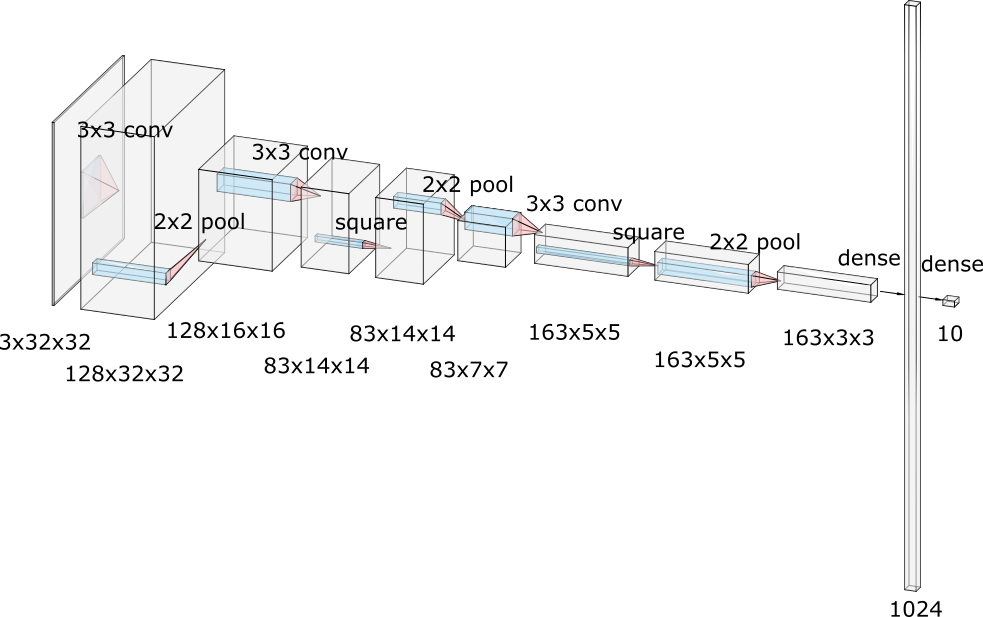}\caption{The structure of the network used for CIFAR classification.\label{fig:cifar-full}}
\end{figure}

\section{CalTech-101}

Table~\ref{tab:CalTech-data-representation} shows the different
data representations when using the method proposed for private inference
on the CalTech-101 dataset using deep representations. \footnote{In Table~\ref{tab:CalTech-data-representation} we use the terminology
	of dense vectors also in the first stage of applying Alex-Net before
	the encryption. }

\begin{table}
	\centering{}%
	\begin{tabular}{|c|c|c|c|}
		\hline 
		\textbf{Layer} & \textbf{Input size} & \textbf{Output format} & \textbf{Description}\tabularnewline
		\hline 
		\hline 
		Preprocess & $200\times300$ & dense & apply convolution layers from Alex-Net \tabularnewline
		\hline 
		Encryption & $4096$ & dense & image is encrypted into $1$ message\tabularnewline
		\hline 
		dense layer & $101$ & sparse & dense-vector row major multiplication\tabularnewline
		\hline 
	\end{tabular}\caption{Data representation changes for CalTech 101 task\label{tab:CalTech-data-representation}}
\end{table}

\end{document}